\def\year{2022}\relax
\documentclass[letterpaper]{article} % DO NOT CHANGE THIS
\usepackage{aaai22}  % DO NOT CHANGE THIS
\usepackage{times}  % DO NOT CHANGE THIS
\usepackage{helvet}  % DO NOT CHANGE THIS
\usepackage{courier}  % DO NOT CHANGE THIS
\usepackage[hyphens]{url}  % DO NOT CHANGE THIS
\usepackage{graphicx} % DO NOT CHANGE THIS
\usepackage{natbib}  % DO NOT CHANGE THIS AND DO NOT ADD ANY OPTIONS TO IT
\usepackage{caption} % DO NOT CHANGE THIS AND DO NOT ADD ANY OPTIONS TO IT
\DeclareCaptionStyle{ruled}{labelfont=normalfont,labelsep=colon,strut=off} % DO NOT CHANGE THIS

\usepackage{enumitem}
\usepackage{multirow}
\usepackage{amssymb}
\usepackage{algorithm}
\usepackage{algorithmic}
\usepackage{newfloat}
\usepackage{listings}
\floatstyle{ruled}
\newfloat{listing}{tb}{lst}{}
\floatname{listing}{Listing}
\title{ARTS: Eliminating Inconsistency between Text Detection and Recognition with Auto-Rectification Text Spotter}

\author{
    Humen Zhong \textsuperscript{\rm 1}\thanks{Zhong did this work during an internship in Alibaba DAMO Academy.},
    Jun Tang \textsuperscript{\rm 2},
    Wenhai Wang \textsuperscript{\rm 1},
    Zhibo Yang \textsuperscript{\rm 2},
    Cong Yao \textsuperscript{\rm 2},
    Tong Lu \textsuperscript{\rm 1}\thanks{Corresponding author}
}

\affiliations{
    \textsuperscript{\rm 1} National Key Lab for Novel Software Technology, Nanjing University\\
    \textsuperscript{\rm 2} Alibaba Group\\
    \{zhonghumen, wangwenhai362\}@smail.nju.edu.cn, xixing.tj@alibaba-inc.com, \{yangzhibo450, yaocong2010\}@gmail.com, lutong@nju.edu.cn
}

\usepackage{bibentry}
\begin{document}

\maketitle

\def\ie{\emph{i.e.}}
\def\eg{\emph{e.g.}}
\def\etc{\emph{etc}}
\def\etal{{\em et al.~}}
\def\vs{\emph{vs.}}

\begin{abstract}
Recent approaches for end-to-end text spotting have achieved promising results. However, most of the current spotters were plagued by the inconsistency problem between text detection and recognition.
In this work, we introduce and prove the existence of the inconsistency problem and analyze it from two aspects:
% In this work, we introduce and analyze the two aspects of the inconsistency problem: 
(1) inconsistency of text recognition features between training and testing, and (2) inconsistency of optimization targets between text detection and recognition.
To solve the aforementioned issues, 
% we first design an Adaptive Training Protocol (ATP) to improve robustness and eliminate the first aspect of the inconsistency problem. Then, 
we propose a differentiable Auto-Rectification Module (ARM) together with a new training strategy to enable propagating recognition loss back into detection branch, so that our detection branch can be jointly optimized by detection and recognition targets, which largely alleviates the inconsistency problem between text detection and recognition. %which largely alleviates the second aspect of the introduced inconsistency problem.
Based on these designs, we present a simple yet robust end-to-end text spotting framework, termed \textbf{A}uto-\textbf{R}ectification \textbf{T}ext \textbf{S}potter (ARTS), to detect and recognize arbitrarily-shaped text in natural scenes.
% we design a differentiable Auto-Rectification Module (ARM) to propagate recognition loss back into detection branch, as well as an Adaptive Sampling Training Protocol (ASTP) to improve the robustness. Based on these, we present a simple end-to-end text spotting framework, termed \textbf{A}uto\textbf{R}ectification \textbf{T}ext \textbf{S}potter (ARTS), to fast detect and recognize arbitrarily-shaped text in natural scenes.
% Compared to previous state of the arts, our ARTS has two merits. (1) Benefiting from the ASTP, we eliminate the first aspect of the inconsistency problem, \ie the inconsistency of text recognition features between training and testing; (2) With the help of ARM, our detection branch can be jointly optimized by detection and recognition targets, largely alleviating the second aspect of the inconsistency problem.
Extensive experiments demonstrate the superiority of our method. In particular, our ARTS-S achieves 77.1\% end-to-end text spotting F-measure on Total-Text at a competitive speed of 10.5 FPS, which significantly outperforms previous methods in both accuracy and inference speed.

\end{abstract}

\section{Introduction}
% Scene text spotting in the wild has attracted increasing attention in recent years due to its numerous applications, such as image understanding, unmanned supermarket, and AR Translator.
Scene text spotting has witnessed remarkable progress and achieved promising results~\cite{2018cvpr_liu_fots,2018eccv_lyu_masktextspotterv1,2020cvpr_liu_abcnet,2020eccv_liao_masktextspotterv3} in recent years.
% Text spotting methods usually follow a locate -and-recognize paradigm. For segmentation-based method, it first segments text area, and then extract features of text through a post-process step to fulfill text-spotting. 
However, there is still room for improvement in scene text spotting, due to the inconsistency between text detection and recognition, which involves the following two aspects.
% it is still challenging in terms of accuracy and efficiency when end-to-end detecting and recognizing text lines with arbitrary layouts
% due to some flaws as follows:
% (1) The ``detection quality gap'' between training and inference, and (2) The inability of detection branch to learn from recognition targets.
% due to a major flaw, \ie the inconsistency/gap between detection and recognition.

\begin{figure}[t]
\begin{center}
   % \includegraphics[width=1.0\linewidth]{LaTeX/images/det_branch_cropped.pdf}
   \includegraphics[width=0.95\linewidth]{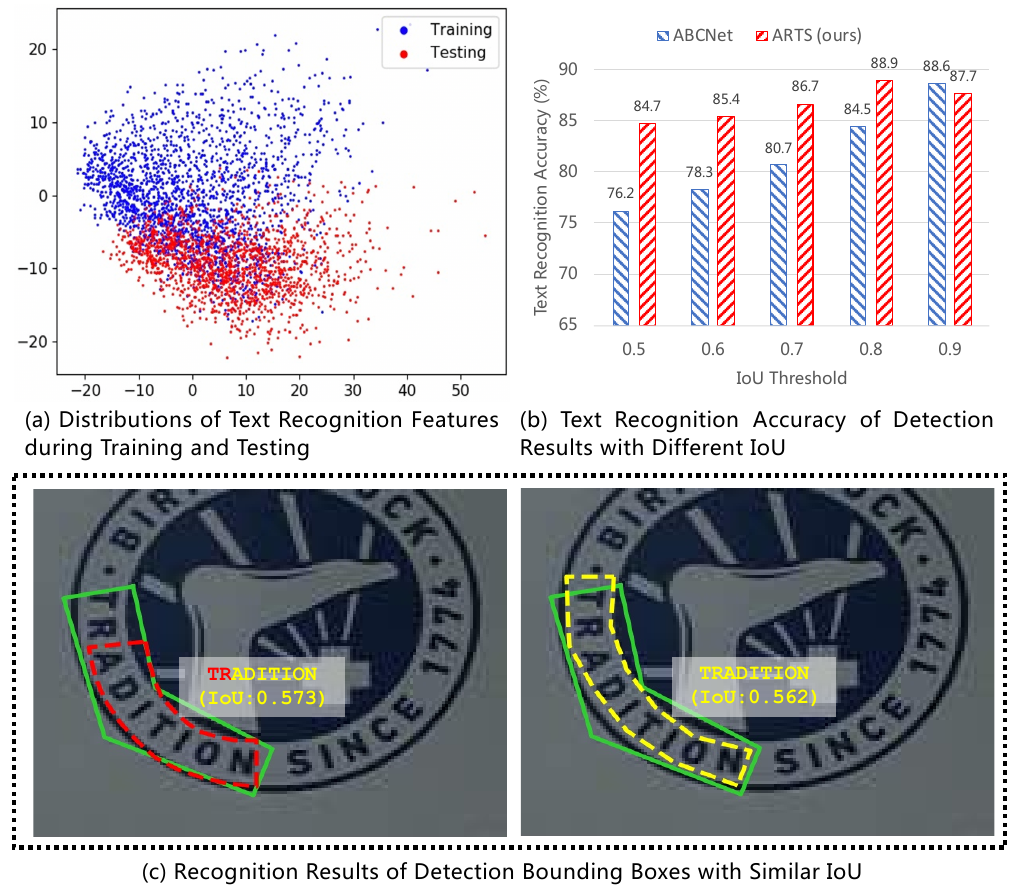}
\end{center}
% \vspace{-0.3cm}
   \caption{Experimental analysis for the proposed inconsistency between text detection and recognition. (a) We collect text recognition features of ABCNet~\cite{2020cvpr_liu_abcnet} during training and testing, to show the inconsistent data distribution; (b) We use different IoU thresholds to select different quality detection results, and evaluate the recognition accuracy of them; (c) An example indicating that detection results with similar IoU can lead to different recognition results.}
\label{fig:motivation}
% \label{fig:onecol}
% \vspace{-0.3cm}
\end{figure}

% （1）识别特征的不一致性，结合fig1的(a)(b)说；（2）检测和识别优化目标的不一致性：检测是iou越高越好，但是同等的iou特征，提取到的识别特征不同。比如iou 0.5的框，可能会漏字符。（结合Fig1的（c）（d）说）
% Specifically, the inconsistency (1) the inconsistency between text recognition features during training and inference and (2)  the inconsistency between the optimization target of text detection and recognition.
% the inability of text detection branch to learn from recognition information.
\textit{The first aspect is the inconsistency of text recognition features during training and testing.}
% during training, previous methods tended to directly use high-quality ground-truth annotations for feature extraction and produced good text features. But during inference, predicted polygons can hardly achieve the quality level of ground-truth annotations and thus will produce worse text features for recognition. 
% This gap makes the recognition head less robust, thus tiny disturbances or mistakes of predicted detection results may easily lead to a false recognition result, bringing a fragile text spotter.
Most existing methods~\cite{2018cvpr_liu_fots,2018eccv_lyu_masktextspotterv1,2020cvpr_liu_abcnet,2020eccv_liao_masktextspotterv3} extract recognition features based on ground-truth annotations in the training phase and predicted bounding boxes in the testing phase, 
which often leads to inconsistent text recognition feature distributions (see Figure~\ref{fig:motivation}(a)).
% , and the poor robustness of the recognition head when processing detection results with $\mbox{IoU}<0.8$ (see Figure~\ref{fig:motivation}(b) blue columns).

% As for the second aspect, previous methods' detection branches are only supervised by detection targets and are highly independent of information in the recognition losses. So they may predict detection results that are not suitable for the recognition task. We argue that it is not enough to just learn from detection targets for our scene text spotter. To generate detection results that are more accurate and more suitable for the subsequent recognition task, it is also essential for our detection branch to directly learn from recognition targets.
\textit{Second, there is inconsistency between optimization targets of text detection and recognition.} Detection branches in the existing methods~\cite{2018cvpr_liu_fots,2018eccv_lyu_masktextspotterv1,2020cvpr_liu_abcnet,2020eccv_liao_masktextspotterv3} are typically optimized to learn high-IoU text detection results.
However, the detection result with high IoU is not always suitable for the recognition task.
% whai: 把图1（c）的右边的结果弄成0.4的
As shown in Figure~\ref{fig:motivation}(c), the bounding box with $\mbox{IoU} > 0.5$ results in the false recognition result, while the bounding box with lower IoU yields the correct result.

Due to the introduced inconsistency between text detection and recognition, previous methods (\eg, ABCNet) suffer a significant performance drop (see Figure~\ref{fig:motivation}(b)) when the IoU of detection results are lower than 0.8, which indicates that although some detection results are considered as ``correct'' under the detection evaluation protocol (\eg, $\mbox{IoU} > 0.5$), these detection results may not be suitable for text recognition (see Figure~\ref{fig:motivation}(c)).
% Detection branches of previous methods are optimized by detection targets only. Since they never learn from recognition information, it is hard for them to produce detection results that are suitable for the following recognition task. Thus, we argue it is important to alleviate the inconsistency by making our detection branch optimized by not only detection targets but also recognition targets. 

To address the aforementioned problems, we propose a new arbitrarily-shaped text spotting framework, termed Auto-Rectification Text Spotter (ARTS), which bridges the inconsistency between text detection and recognition.
%
% as illustraed in Figure~\ref{fig:overall_pipeline}.
We carefully design three modules for ARTS, which include: (1) a rectification control points detection (RCPD) branch to detect arbitrarily-shaped text lines; (2) a differentiable feature extractor termed auto-rectification module (ARM) for back-propagating text recognition loss to optimize the detection branch; and (3) a lightweight text recognition branch to decode text contents. 
All the modules above complement each other, enabling the proposed ARTS to learn text detection results from both detection loss and recognition loss, which largely alleviates the inconsistency problem between text detection and recognition.
As the red columns shown in Figure~\ref{fig:motivation}(b) and the example shown in Figure~\ref{fig:motivation}(c), our method achieves much better performance especially when the detection results are with lower-quality (IoU $<$ 0.8).

%这个精度分数表示了检测分支生成的检测框对于识别任务的合适程度。

% text recognition features inconsistency
% detection branches can not be optimized by recognition losses
%gap between detection and recognition

\begin{figure}[t]
\begin{center}
   % \includegraphics[width=0.95\linewidth]{LaTeX/images/fps_fscore_cropped_v2.pdf}
   \includegraphics[width=0.85\linewidth]{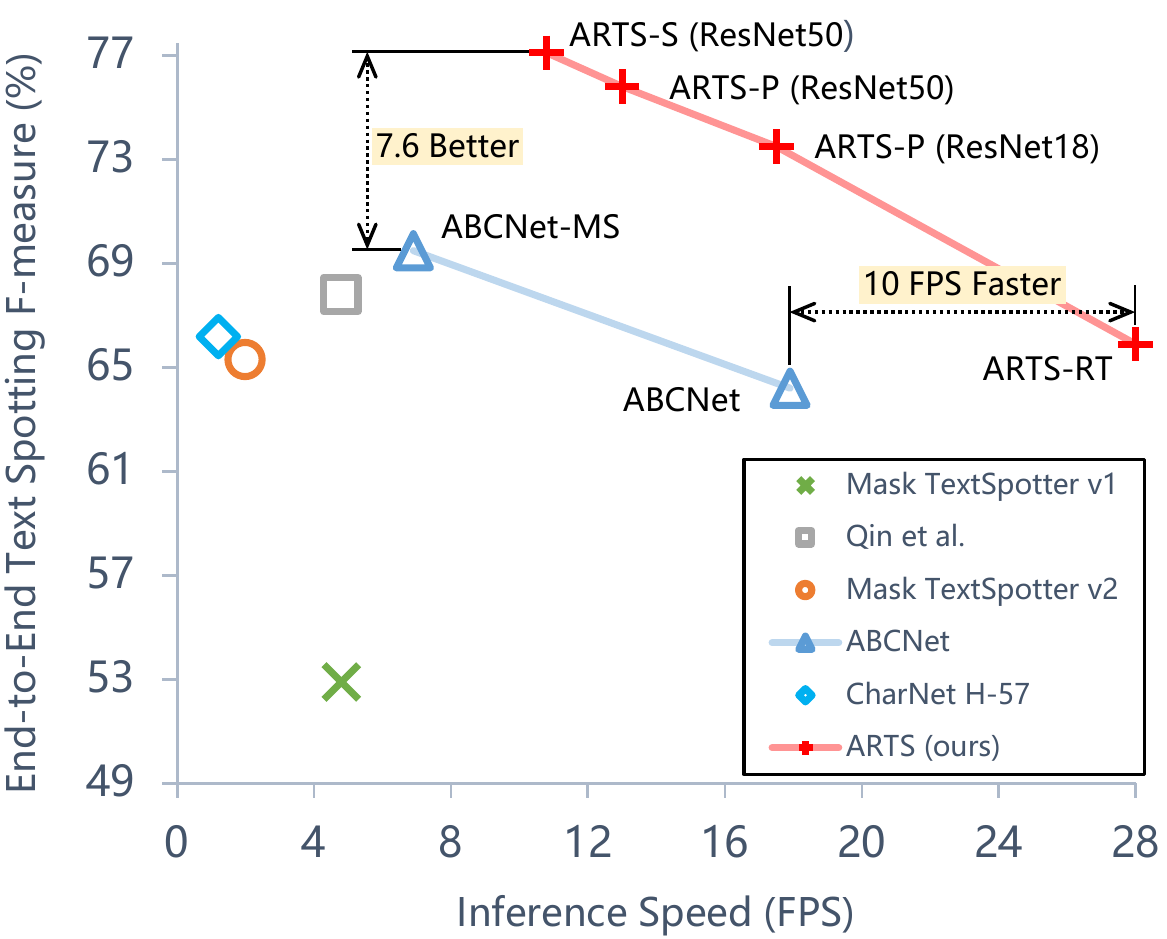}
\end{center}
% \vspace{-0.3cm}
   \caption{Performances of different methods on Total-Text. 
   The proposed ARTS models achieve significantly better trade-off between accuracy and inference speed than previous methods.
   % achieves an end-to-end text spotting F-measure (E2E F-measure) of 77.1\% at a competitive speed (10.5FPS). And ARTS-RT surpasses previous fastest ABCNet by 1.7\% and runs 10FPS faster.
   }
\label{fig:fps_fscore}
% \label{fig:onecol}
% \vspace{-0.3cm}
\end{figure}

\begin{figure*}[t]
\begin{center}
% \fbox{\rule{0pt}{2in} \rule{.9\linewidth}{0pt}}
    \includegraphics[width=0.85\linewidth]{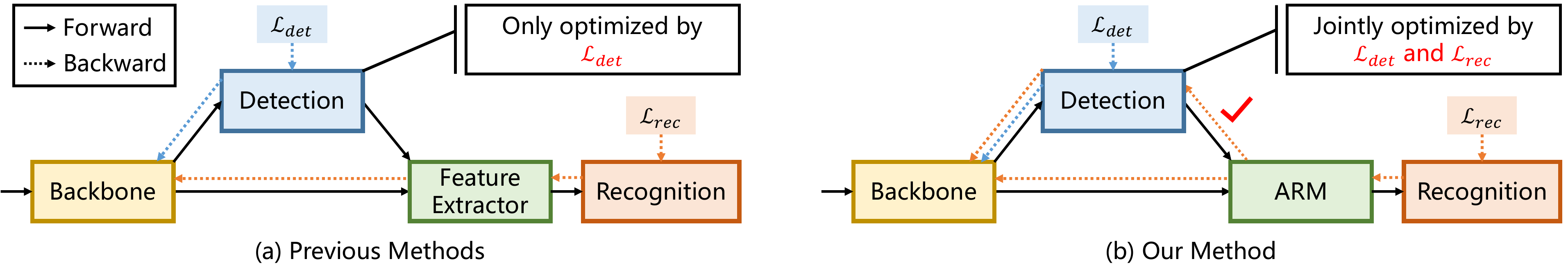}
\end{center}
% \vspace{-0.3cm}
   \caption{Illustration of different pipelines. Previous ``less'' end-to-end methods cannot back-propagate recognition loss to the detection branch, so that their detection branches can only be  optimized by detection targets. While our method allow the loss back-propagation from the recognition branch to the detection branch, and thus our detection branch is jointly optimized by detection and recognition targets.}
\label{fig:diff_pipeline}
% \vspace{-0.3cm}
\end{figure*}

We conduct extensive experiments to further examine the effectiveness of ARTS on three challenging benchmark datasets, including Total-Text~\cite{2017icdar_totaltext}, CTW1500~\cite{2017arxiv_liu_ctw1500} and ICDAR2015~\cite{2015icdar_ic15}. As shown in Figure~\ref{fig:fps_fscore}, our method surpasses prior arts in terms of both accuracy and efficiency. For example, our ARTS-S (ResNet50) achieves an end-to-end text spotting F-measure of 77.1\% on Total-Text, surpassing ABCNet-MS~\cite{2020cvpr_liu_abcnet} by 7.6 points, while keeping a faster inference speed (10.5 FPS \vs\ 6.9 FPS).
Moreover, the real-time version ARTS-RT yields an F-measure of 65.9\% at 28.0 FPS, which is 10 FPS faster and 1.7\% better than the previous fastest ABCNet.

% 这里再强调一下我们的方法变得更robust了，结合第一张图的（b）。

% 这里还能说是previous fastest吗？pan++和pgnet

% By changing our text recognition head's structure, we can even achieve a better performance with an F-Score of ? and a inference speed of ?.

Our main contributions are listed as follows:
% \begin{itemize}

(1) We systematically analyze the inconsistency between text detection and recognition, and propose a new text spotting framework, termed ARTS, to address this problem. To our knowledge, our method is the first work to study and tackle the inconsistency problem in text spotting.
% \item We introduce and analyze the performance effect of the ``quality gap problem'' between predicted detection results and ground-truth annotations. And in the meanwhile, we further argue that it is essential to let detection branch learn from recognition targets for more suitable detection results.

(2) We design a differentiable module named ARM to bridge the gap between text detection and recognition branches, so that recognition loss can be back-propagated to optimize the detection results, helping detection branch to predict more accurate and more suitable detection results for text recognition.

(3) The proposed ARTS achieves state-of-the-art performance in terms of both accuracy and efficiency. Extensive experiments demonstrate the superiority of our models. Notably, ARTS-S (ResNet50) yields 77.1\% end-to-end text spotting F-measure at 10.5 FPS on Total-Text, which is significantly better and faster than previous state-of-the-art methods.
% \end{itemize}

\section{Related Work}
% \subsection{End-to-End Text Spotting} 
Existing text spotting methods can be roughly summarized into the following two categories:

% Early works separate the entire pipeline into two steps, i.e. detecting text instances first and then cropping the text region images and seed them into a trained text recognizer. However, two separate networks can not utilize the complementary information between detection and recognition and will lead to a sub-optimal performance. In recent years, researchers have been investing great efforts into end-to-end text spotting and have achieved promising results.

\subsubsection{Regular 
% End-to-End Scene
Text Spotters} are usually designed to process horizontal or multi-oriented scene text. DeepTextSpotter~\cite{2017iccv_busta_deeptextspotter} used  RPN to generate rotated proposals and extracted text features for its recognizer with bilinear sampling. FOTS~\cite{2018cvpr_liu_fots} adopted a one-stage text detector to produce rotated rectangular bounding boxes and used RoIRotate to extract text features for the following recognizer. 
% But these methods are not able to spot irregular texts.
He \etal\cite{2018cvpr_he_e2etextspotter} also developed a similar framework whose recognition head was implemented by an attention-based decoder.
Though these methods have achieved promising results on standard benchmarks (\eg, ICDAR 2015~\cite{2015icdar_ic15}), they failed to spot texts with arbitrary shapes.
% \whai{A similar framework is also developed by He \etal\cite{he2018end}, whose
% recognition head is implemented by an attention-based sequence-to-
% sequence decoder.
% Although these methods have achieved good
% performance across straight text benchmarks (e.g., IC15~\cite{2015icdar_ic15}), they fail to detect and recognize text lines of
% curved shapes, as shown in Figure \ref{fig:diff_repr} (a).}

\subsubsection{
% Arbitrary-Shaped 
Arbitrarily-Shaped 
% End-to-End Scene 
Text Spotters} 
% on the contrary 
are designed for spotting
% to detect and recognize
texts with irregular layouts.
% 下面这一句需要在这里重复一遍吗？
% As shown in Figure \ref{fig:diff_repr}, there have been various text representations designed for irregular scene text, including the bound-box-mask representation~\cite{2018eccv_lyu_masktextspotterv1,2019pami_liao_masktextspotterv2,2019iccv_qin_unconstrained}, the kernel-based representation~\cite{2019cvpr_wang_psenet,wang2019efficient,2019iccv_feng_textdragonv1,song2020tk}, and Bezier representation~\cite{2020cvpr_liu_abcnet}.
%
TextDragon~\cite{2019iccv_feng_textdragonv1} developed a bottom-up framework to combine features extracted from multiple text segments by RoISlide. Mask TextSpotter v1/v2~\cite{2018eccv_lyu_masktextspotterv1,2019pami_liao_masktextspotterv2} and Qin \etal~\cite{2019iccv_qin_unconstrained} were based on Mask R-CNN~\cite{2017iccv_he_maskrcnn} and extracted recognition features through RoIAlign or RoIMasking.
% Boundary~\cite{2020aaai_wang_boundary} adopted a multi-stage anchor-based model to detect arbitrarily-shaped text lines.
% \whai{TextDragon~\cite{2019iccv_feng_textdragonv1} is 
% % proposed to spot arbitrarily-shaped text in 
% a bottom-up method that 
% % fused the extracted
% combines the features extracted from multiple text segments by RoISlide.}
% \whai{Mask TextSpotter v1/v2~\cite{2018eccv_lyu_masktextspotterv1}\cite{2019pami_liao_masktextspotterv2} and Qin \etal \cite{2019iccv_qin_unconstrained} are text spotters based on Mask R-CNN~\cite{2017iccv_he_maskrcnn}, which detect scene text via a two-stage detector.}
% proposed to spot irregular text with a two-stage Mask R-CNN style text detector and extracted aligned features for recognition by RoIAlign or RoIMasking.
% CharNet~\cite{2019iccv_xing_charnet} used a segmentation framework to concurrently predict word-level and character-level text instances.
Wang \etal \cite{2020aaai_wang_boundary} utilized a multi-stage anchor-based method to first generate axis-aligned rectangular proposals, then regress their angles to produce rotated rectangular proposals and finally regress boundary points on top of the rotated rectangular proposals.
ABCNet~\cite{2020cvpr_liu_abcnet} proposed to use parametric bezier control points as the representation for arbitrary-shaped text instances to extract smooth text feature.

\subsubsection{Comparison with Similar Works.}
% \begin{itemize}

% \noindent\textbf{Comparison with Boundary~\cite{2020aaai_wang_boundary}.}
Boundary~\cite{2020aaai_wang_boundary} adopted a three-stage anchor-based detector as its detection branch and cannot back-propagate recognition losses to the first two detection stages. Differently, our method adopt a much simpler one-stage anchor-free pipeline as our detection branch and our detection branch can be jointly optimized by detection and recognition targets as shown in Figure~\ref{fig:diff_pipeline}(b), 
largely alleviating the inconsistency between text detection and recognition.

% \noindent\textbf{Comparision with ABCNet~\cite{2020cvpr_liu_abcnet}.}
ABCNet~\cite{2020cvpr_liu_abcnet} adopted BezierAlign for feature extraction. But we argue that BezierAlign also cannot propagate recognition loss back to the detection branch, causing the inconsistency between text detection and recognition. In our work, we use a different training strategy, \ie, using predicted instead of ground-truth polygons to extract text features for recognition task during training, making it possible for our further improvement, \ie, we use our ARM to extract text features and further enable loss back-propagation from recognition towards detection branch, which is hard to achieve by ABCNet's BezierAlign.

\begin{figure*}
\begin{center}
% \fbox{\rule{0pt}{2in} \rule{.9\linewidth}{0pt}}
    % \includegraphics[width=0.95\linewidth]{LaTeX/images/overall_pipeline_v1.pdf}
    \includegraphics[width=0.95\linewidth]{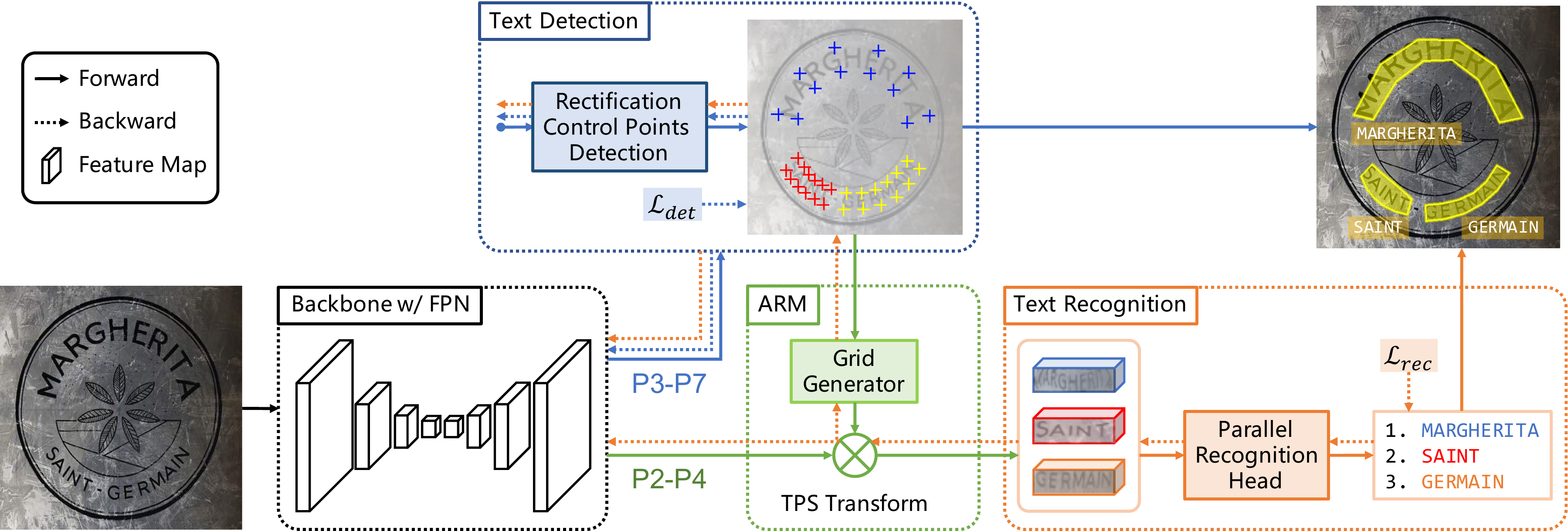}
\end{center}
\caption{Illustration of the overall architecture. Note that, rectification control points detection is densely conducted on every level of FPN output features, here we just depict one of them for clarity.}
\label{fig:overall_pipeline}
\end{figure*}

\section{Methodology}
\subsection{Overall Architecture}
ARTS is an efficient and accurate end-to-end framework for detecting and recognizing text lines with arbitrary shapes. The overall architecture of ARTS is presented in Figure~\ref{fig:overall_pipeline}, which consists of three components: (1) a Rectification Control Points Detection head (RCPD) to detect and predict control points for each text line, (2) a differentiable Auto-Rectification Module (ARM) to rectify curved text features into aligned ones and allow loss back-propagation from recognition to detection branch, and (3) a text recognition branch to decode text contents from extracted features.

In the forward phase, we first feed the input image to the backbone network and output the shared feature maps. Secondly, on top of the feature maps, RCPD predicts the text location and the rectification control points. Thirdly, these predicted rectification control points will be sent to ARM for rectifying and extracting text features. Finally, the aligned features are fed into the text recognition head to obtain the final text contents.

% 这里要不要强调一下反向传播耽时候的传播路径？强调可以向检测分支传播？

During training, we use the joint loss of detection loss $\mathcal{L}_{det}$ and recognition loss $\mathcal{L}_{rec}$ to optimize our model. Different from previous methods~\cite{2018cvpr_liu_fots,2018eccv_lyu_masktextspotterv1,2019pami_liao_masktextspotterv2,2020cvpr_liu_abcnet} whose detection branches are only supervised by loss function $\mathcal{L}_{det}$, our detection branch is jointly optimized by detection and recognition targets with loss functions $\mathcal{L}_{det}$ and $\mathcal{L}_{rec}$. Besides, unlike previous methods who tended to directly use ground-truth annotations for feature extraction during training, our method adopts a new training strategy to use predicted detection results instead.
Concretely, we define the central region of a text instance as positive pixels, and evenly sample $n_{text}$ pixels from all positive pixels. Then, we use the $n_{text}$ groups of predicted control points of these sampled pixels and send them to our ARM to get $n_{text}$ text recognition features, which will be fed into recognition branch to train our text recognition branch. Here, $n_{text}$ is set to 64 by default.

\subsection{Rectification Control Points Detection}\label{RCPD}
\begin{figure}[t]
\begin{center}
   % \includegraphics[width=1.0\linewidth]{LaTeX/images/det_branch_cropped.pdf}
   \includegraphics[width=0.75\linewidth]{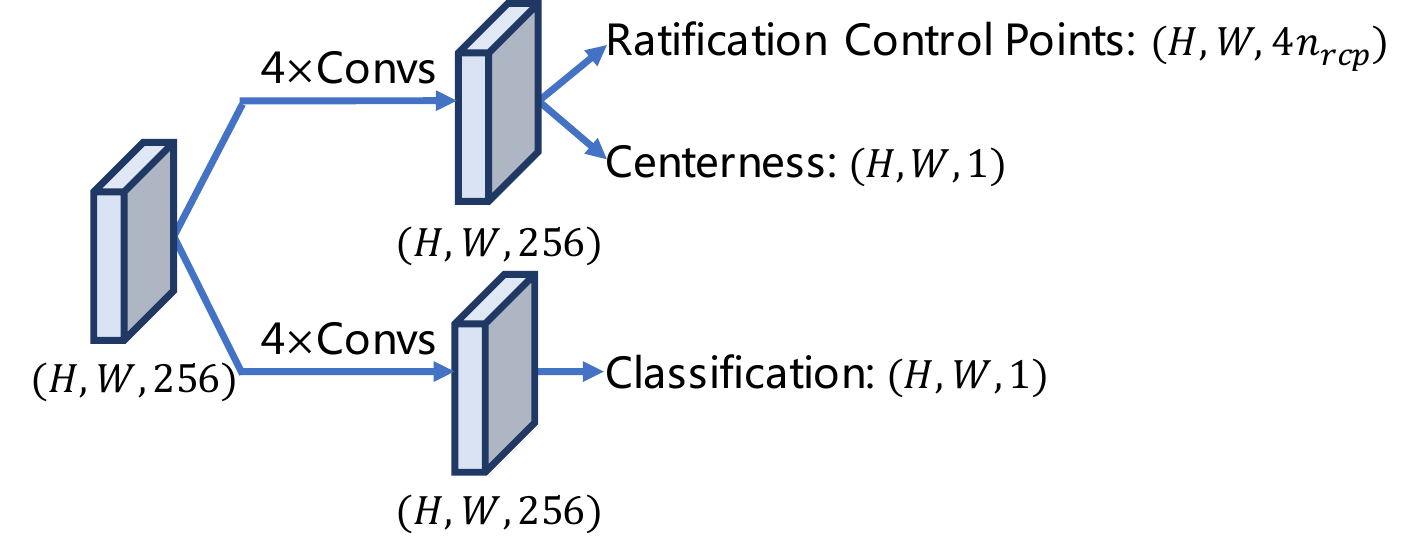}
\end{center}
% \vspace{-0.3cm}
   \caption{Detailed structure of detection branch.}
\label{fig:det_branch}
% \label{fig:onecol}
% \vspace{-0.3cm}
\end{figure}

As presented in Figure~\ref{fig:det_branch}, we adopt a one-stage anchor-free framework as our detection branch to densely regress rectification control points for all text lines.
% As presented in Fig.\ref{fig:det_branch}, we densely predict rectification control points for all text lines. 
For each text line, we sample the central region as positive pixels and regress offsets from the pixel towards the control points of this text line. 
The size of the regression result is $(H/S, W/S, 4 \times n_{rcp})$, where $n_{rcp}$ means the number of control points for each side, $S$ denotes the downsampling scale to the input image, while $H$ and $W$ are the height and width of the feature map, respectively.

\subsubsection{Ground-Truth Generation of RCPD.}
\begin{figure}[t]
\begin{center}
   \includegraphics[width=1.0\linewidth]{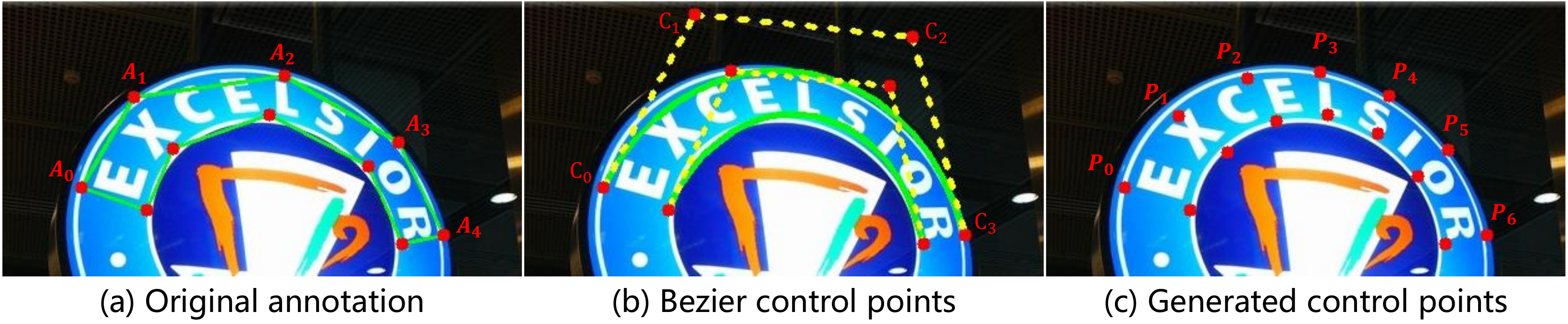}
\end{center}
% \vspace{-0.3cm}
   \caption{Target generation process for rectification control points.}
\label{fig:tps_target_generation}
% \label{fig:onecol}
% \vspace{-0.3cm}
\end{figure}

We do not directly use the annotations as our ground-truth targets because the annotations provided by the dataset are not accurate enough for extracting high-quality text features.
As depicted in Figure~\ref{fig:tps_target_generation}, 
we recalculate the control points targets by first fitting cubic bezier curves, and then uniformly sample $n_{rcp}$ points according to the following equation:
% for a text instance which is \textit{unevenly} annotated
% with several boundary points $A_i$ like in Total-Text~\cite{2017icdar_totaltext} or SCUT-CTW1500~\cite{2017arxiv_liu_ctw1500}, we first fit cubic bezier curves to match its top and bottom boundaries respectively. Then, we uniformly sample $n_{rcp}$ points based on the Bezier control points according to the following equation:
\vspace{-0.2cm}
\begin{equation}
\vspace{-0.2cm}
    P_k=\sum_{i=0}^{n}C_iB_{i,n}(\frac{k}{n_{rcp}})\\
% \vspace{-0.1cm}
\end{equation}
where $P_k$ indicates the \textit{k}-th sampled control points, $C_i$ indicates the \textit{i}-th bezier control points and $n_{rcp}$ is a hyper-parameter which determines how many rectification control points do we sample on each side of text. $B_{i,n}$ represents the Bernstein basis polynomials and is formulated as follows:
% \begin{equation}
% % \vspace{-0.1cm}
%     B_{i,n}(t) = \tbinom{n}{i}t^i{(1-t)}^{n-i}, i=0,...,n
% \end{equation}
% where $\tbinom{n}{i}$ is the binomial coefficient.
\begin{equation}
    B_{i,n}(t) = C_{n}^{i}t^i{(1-t)}^{n-i}
\end{equation}
where $C_{n}^{i}$ is the binomial coefficient.
% \begin{equation}
%     B_{i,n}(t) = \dbinom{n}{i}t^i{(1-t)}^{n-i}
% \end{equation}
% where $C^{i}_{n}$ is the binomial coefficient.

The sampled $n_{rcp}$ points are defined as the rectification control points for this text instance, and are used for generating the training target. Concretely, for a positive pixel at position $(x,y)$, we generate the offset target as follows:
\begin{equation}\label{eq:target}
    \Delta_{k,x}=P_{k,x}-x,\ \ \ \ 
    \Delta_{k,y}=P_{k,y}-y
    % O_{target}(k,x,y)=RCP_k-(x,y)
\end{equation}
where $P_{k,x}$ and $P_{k,y}$ mean coordinates of the \textit{k}-th control point, while $\Delta_{k,x}$ and $\Delta_{k,y}$ denote the target offset towards the \textit{k}-th control point.

% It is worth to be mentioned that, our representation reaches a special case where the network will directly predict the 4 corner points just like what \cite{2017cvpr_zhou_east, 2018cvpr_liu_fots} did when $n_{rcp}$ is set 2.
% Compared with Boundary~\cite{2020aaai_wang_boundary} that adopt a multi-stage anchor-based pipeline to detect text instances and cannot propagate recognition loss back to the first two detection stages, we adopt a much simpler single-stage detection pipeline and our detection branch can be jointly optimized by detection and recognition targets as shown in Figure~\ref{fig:diff_pipeline}. 
% Benefiting from these advantages, our method surpasses Boundary by a large margin (77.1\% \vs 65.0\%) while keeping a competitive speed (10.5 FPS).

\subsection{Auto-Rectification Module}
% In many text recognition methods, rectification is an essential step to rectify curved text features into aligned features and avoid background noise. However, rectification module in these methods can only handle one text instance in a image. 

% Inspired by this, we design an Auto-Rectification Module (ARM) to rectify arbitrarily-shaped text features of all texts to aligned ones for the subsequent recognition task.

% ARM is designed
Previous methods tended to use RoIAlign operator or its variants for feature extraction. However, these operators can only back-propagate recognition loss into the shared backbone but not into the detection branch. Thus their detection branches are supervised by detection targets only and are highly independent of recognition information. These detection branches cannot learn from recognition targets and thus cannot produce detection results that are suitable for text recognition, leading to inconsistency between text detection and recognition.

We propose to design a new feature extractor named Auto-Rectification Module (ARM) to eliminate the inconsistency.
ARM receives $n_{text}$ groups of predicted rectification control points for $n_{text}$ text instances, and outputs $n_{text}$ aligned text features for all text instances. Our ARM is implemented mainly based on a differentiable Spatial Transform Network (STN)~\cite{2015nips_stn}. Note that we further upgrade the original version so that it can handle the situation where there are multiple text instances in the same image.
Due to page limit, detailed mathematical formulation will be provided in supplementary materials, and we refer readers to ~\cite{2015nips_stn} for more detailed information about STN.
Compared with previous methods~\cite{2020cvpr_liu_abcnet,2018eccv_lyu_masktextspotterv1,2019pami_liao_masktextspotterv2,2018cvpr_liu_fots}, our proposed module has the following differentiability advantage.

\subsubsection{Differentiability from Recognition to Detection.}
Previous end-to-end methods like~\cite{2020cvpr_liu_abcnet,2018cvpr_liu_fots} only share backbone features but often lack the ability of back-propagating recognition loss into detection branch.
% That is to say, they directly use ground-truth annotations to extract features for training their recognition branch so that losses from recognition can not help detection branch to do better.
% The two branches just work alone without more cooperation. 
We argue that it is of vital importance for our RCPD head to learn from recognition losses for producing better detection results. 
So in our framework, we propose to use ARM, which is completely differentiable to enable loss back-propagation from recognition to our RCPD head. As a result, our RCPD head, which will be jointly optimized by detection and recognition targets, can predict more suitable results for the subsequent recognition task. Extensive results also verify our argument that learning from recognition losses can help the entire network achieve global-optimal and obtain better performance in end-to-end text spotting metric.

% \subsubsection{Naturally fitted with text recognition.}

% \subsubsection{Adaptive Training Protocol.}
% In addition to the proposed ARM, we need one more thing, \ie a new training strategy, to enable the loss back-propagation from recognition to detection.

% Concretely, during training process, we randomly sample a batch of positive pixels and obtain their predicted rectification control points for each pixel. Then we use our proposed Auto-Rectification Module to rectify these curved text features into aligned ones and send them into our recognition branch. 
% Note that, we are using control points predicted by the RCPD head during training \emph{instead of} ground-truth control points, which allow the network to back-propagate recognition loss into detection branch and help achieve global optimal.

\subsection{Recognition Branch}
% Benefiting from our differentiable Auto-Rectification Module, curved text features can be automatically rectified into aligned ones, making it possible for a light-weight recognition branch.
% Specifically, our recognition branch contains a feature extractor, a sequence modeling module and a decoder.
To validate the effectiveness and robustness of our spotting framework, we adopt two different recognizers, \ie, Parallel Recognizer and Serial Recognizer. Both the recognizers have the same feature extractor, but differ in their sequence modeling modules and decoders.
The detailed structure of our recognition branch can be seen in Table~\ref{tab:rec_structure}.

\begin{table}[ht!]
% \vspace{-1mm}
\small
\centering
\begin{tabular}{|c|c|c|c|}\hline
\multicolumn{2}{|c|}{Layers} & Parameters & Output Size\\ \hline
\multicolumn{2}{|c|}{Conv layers $\times$ 2} & 3, 1, 1 & (n, 256, h, w) \\
\multicolumn{2}{|c|}{Conv layers $\times$ 1} & 3, (2,1), 1 & (n, 256, h/2, w) \\
\multicolumn{2}{|c|}{Conv layers $\times$ 2} & 3, 1, 1 & (n, 256, h/2, w) \\
\multicolumn{2}{|c|}{Conv layers $\times$ 1} & 3, (2,1), 1 & (n, 256, h/4, w) \\
\multicolumn{2}{|c|}{Avg \& Permute} & - & (w, n, 256)\\
\hline
BiLSTM & Self-Attn & - & (w, n, 256) \\
\hline
Serial & Parallel & - & (n, len, n\_class)\\
\hline
\end{tabular}
\caption{Detailed structures of our recognition branch. "Serial" is for Serial decoder and "Parallel" is for Parallel decoder. "Parameters" is for kernel-size, stride and padding.}\label{tab:rec_structure}
% \vspace{-3mm}
\end{table}

\subsection{Loss Function}

The overall loss function of our model consists of two parts: (1) detection loss ${\mathcal{L}}_{det}$ and (2) recognition loss ${\mathcal{L}}_{rec}$. It is defined as follows: 
\begin{equation}
    {\mathcal{L}}=\lambda_{det}{\mathcal{L}}_{det}+\lambda_{rec}{\mathcal{L}}_{rec}
\end{equation}

The detection loss function ${\mathcal{L}}_{det}$ is a multi-task loss function which can be defined as Eqn~\ref{eqn:det}.
\begin{equation}\label{eqn:det}
    {\mathcal{L}}_{det}={\mathcal{L}}_{cls}+{\mathcal{L}}_{ctr}+\lambda_{rcp}{\mathcal{L}}_{rcp}
\end{equation}
where ${\mathcal{L}}_{cls}$ and ${\mathcal{L}}_{ctr}$ are for classification and centerness prediction, respectively, which is similar to loss function used in ~\cite{2019iccv_tian_fcos}. ${\mathcal{L}}_{rcp}$ is the loss function of our RCPD head, which is implemented by Smooth L1 loss~\cite{2015iccv_girshick_fastrcnn} and is formulated as follows:
\begin{equation}
    {\mathcal{L}}_{rcp}=\mbox{Smooth}_{L_1}(\Delta_{pred}, \Delta_{target})
\end{equation}
where $\Delta_{pred}$ and $\Delta_{target}$ are the predicted offsets and target offsets of rectification control points defined in Eqn~\ref{eq:target}, respectively. Here
$\lambda_{rcp}$ is used to balance the importance and is set to 0.2 by default in our experiments.
The recognition loss function ${\mathcal{L}}_{rec}$ is for optimizing the recognition branch and follows a similar loss function used in ~\cite{2018pami_shi_aster, 2019iccv_baek_whatiswrong}.

\section{Experiments}
\subsection{Datasets}
Our training process is divided into two phases, that is, pretraining and finetuning. During pretraining, we use a mixed dataset consisting of SynthText150k~\cite{2020cvpr_liu_abcnet}, Total-Text~\cite{2017icdar_totaltext} and MLT~\cite{2017icdar_mlt17}. As for finetuning, we finetune our network on target datasets, \ie, Total-Text, CTW1500 and ICDAR2015, respectively.

\subsection{Implementation Details}
\subsubsection{Network Details.} The backbone of our network follows a common setting as most of the previous papers~\cite{2020cvpr_liu_abcnet, 2020aaai_wang_boundary, 2018eccv_lyu_masktextspotterv1, 2019pami_liao_masktextspotterv2}, \ie, ResNet-50~\cite{2016cvpr_he_resnet} together
with a Feature Pyramid Network (FPN)~\cite{2017cvpr_lin_fpn}. Following the settings of previous papers, for detection branch, we conduct dense prediction on 5 feature maps with 1/8, 1/16, 1/32, 1/64, 1/128 resolution of the input image while for ARM and the subsequent recognition, we use 3 feature maps with 1/4, 1/8, 1/16 resolution.

\subsubsection{Training Details.} We train our model with a batchsize of 8, using Stochastic Gradient Descent (SGD) with momentum of 0.9. The maximum iteration of pretraining is 260K and the initial learning rate is set to 0.02, which decays to a tenth at $160K^{th}$ and $220K^{th}$ iteration. As for finetuning, the maximum iteration is 10K for Total-Text and IC15, which decays to a tenth at $7K^{th}$ and $9K^{th}$ iteration and 130K for CTW1500, which decays to a tenth at $80K^{th}$ iteration. Following prior arts, we adopt widely-used data augmentation strategies: (1) instance aware random cropping, (2) random scaling with shorter side randomly chosen from 640 to 896, and (3) random rotation with angle randomly chosen from [${-45}^{\circ}$, ${+45}^{\circ}$].

\subsubsection{Inference Details.}
% During inference, we first detect irregular text instances with our detection branch and produce a group of control points for each text instance. Then, we use our ARM to rectify curved text features into aligned ones. At last, we send these aligned features into our recognition branch and decode the text contents.
We resize the shorter side of the input image to 1000 for Total-Text, 800 for CTW1500 and 1000 for ICDAR2015. We use NMS to filter out overlapped predictions and the threshold is set to 0.5. All the results are tested with batchsize of 1 using one Tesla V100 GPU.
For the best detection metric, we use a confidence threshold of 0.4 to filter out texts with low detection scores. And for the best end-to-end metric on Total-Text, we use a recognition threshold of 0.9 (for None) or 0.7 (for Full) to filter out texts with low recognition scores, which can be simply calculated by averaging scores for all characters.

\subsection{Comparisons with State-of-the-Art methods}
\subsubsection{Arbitrarily-Shaped Text Spotting.}

\begin{table*}[ht!]
% \vspace{-1mm}
\centering
% \caption{Quantitative results on Total-Text~\cite{2017icdar_totaltext}. "None" and "Full" indicates results with no lexicon and full lexicon respectively. ARTS-P and ARTS-S indicates using parallel and serial decoder respectively. "RT" means a real-time R18 version which shrink the size of input image to 640 for short side. $\dagger$ indicates using private data for training.
% % * indicates the best detection \textbf{and} end-to-end f-score that we can simultaneously achieve (see Sec.~\ref{test_detail} for details).
% }\label{tab:results_totaltext}
\small
\begin{tabular}{|c|c|c|c|c|c|c|c|c|}\hline
\multirow{2}{*}{Method} & \multirow{2}{*}{Venue} & \multirow{2}{*}{Backbone} & \multicolumn{3}{c|}{Detection} & \multicolumn{2}{c|}{End-to-End} & \multirow{2}{*}{FPS} \\
\cline{4-8} & & & Precision & Recall & F-measure & None & Full & \\ \hline
% EAST~\cite{2017cvpr_zhou_east} & CVPR'17 & 50.0 & 36.2 & 42.0 & - & - & -\\
% TextSnake~\cite{2018eccv_long_textsnake} & ECCV'18 & \\
% ATRR[] & CVPR'19 & 80.9 & 76.2 & 78.5 & - & - & 10.0\\
% LOMO[] & CVPR'19 & 88.6 & 75.7 & 81.6 & - & - & -\\
% PSENet~\cite{2019cvpr_wang_psenet} & CVPR'19 & 84.0 & 78.0 & 80.9 & - & - & 3.9\\ 
% \hline
MaskTextSpotterv1~\cite{2018eccv_lyu_masktextspotterv1} & ECCV'18 & ResNet50 & 69.0 & 55.0 & 61.3 & 52.9 & 71.8 & 4.8\\
MaskTextSpotterv2~\cite{2019pami_liao_masktextspotterv2} & PAMI'19 & ResNet50 & 88.3 & 82.4 & 85.2 & 65.3 & 77.4 & 2.0 \\
TextDragon~\cite{2019iccv_feng_textdragonv1} & ICCV'19 & VGG16 & 85.6 & 75.7 & 80.3 & 44.8 & 74.8 & -\\
Unconstrained $\dagger$~\cite{2019iccv_qin_unconstrained} & ICCV'19 & ResNet50 & 83.3 & 83.4 & 83.3 & 67.8 & - & 4.8 \\
CharNet~\cite{2019iccv_xing_charnet} & ICCV'19 & Hourglass57 & \textbf{89.9} & 81.7 & 85.6 & 66.6 & - & 1.2\\
ABCNet~\cite{2020cvpr_liu_abcnet} & CVPR'20 & ResNet50 & - & - & - & 64.2 & 75.7 & 17.9\\
ABCNet MS~\cite{2020cvpr_liu_abcnet} & CVPR'20 & ResNet50 & - & - & - & 69.5 & 78.4 & 6.9\\
Boundary~\cite{2020aaai_wang_boundary} & AAAI'20 & ResNet50 & 88.9 & \textbf{85.0} & \textbf{87.0} & 65.0 & 76.1 & -\\
TextPerceptron~\cite{2020aaai_qiao_textperceptron} & AAAI'20 & ResNet50 & 88.8 & 81.8 & 85.2 & 69.7 & 78.3 & -\\
MaskTextSpotterv3~\cite{2020eccv_liao_masktextspotterv3} & ECCV'20 & ResNet50 & - & - & - & 71.2 & 78.4 & -\\ 
\hline
% \textbf{ARTS-RT (ours)} & - & ResNet18 & 87.3 & 77.6 & 82.2 & 70.3 & 80.9 & \textbf{23.8}\\
ARTS-RT (ours) & - & ResNet18 & 86.8 & 74.8 & 80.3 & 65.9 & 78.1 & \textbf{28.0}\\
ARTS-P (ours) & - & ResNet18 & 86.9 & 81.5 & 84.1 & 73.5 & 83.5 & 17.0\\
ARTS-P (ours) & - & ResNet50 & 88.8 & 83.8 & 86.2 & 75.8 & \textbf{85.4} & 13.0 \\
ARTS-S (ours) & - & ResNet50 & 89.3 & 84.0 & 86.5 & \textbf{77.1} & 85.1 & 10.5\\
% \textbf{ARTS-S* (ours)} & - & 92.7 & 79.2 & 85.4 & 76.6 & 85.1 & 10.5\\
\hline
\end{tabular}
\caption{Quantitative results on Total-Text~\cite{2017icdar_totaltext}. "None" and "Full" indicate results with no lexicon and full lexicon, respectively. ARTS-P and ARTS-S indicate using parallel and serial decoder, respectively. "RT" means a real-time R18 version which shrink the size of input image to 640 for short side. $\dagger$ indicates using private data for training.
% * indicates the best detection \textbf{and} end-to-end f-score that we can simultaneously achieve (see Sec.~\ref{test_detail} for details).
}\label{tab:results_totaltext}
% \vspace{-3mm}
\end{table*}

Our network mainly focuses on arbitrarily-shaped text spotting. To verify its effectiveness, we conduct experiments on the challenging Total-Text dataset. We follow the official evaluation protocol in ~\cite{2020cvpr_liu_abcnet} to make a fair comparison.

The results on Total-Text can be seen in Table~\ref{tab:results_totaltext}. Our method outperforms previous state-of-the-art methods by a large margin both in terms of accuracy and efficiency. Concretely, our ARTS-S achieves an outstanding E2E F-measure of 77.1\% without lexicons which surpasses existing methods by +5.9\% (77.1\% \vs\ 71.2\%) with a light-weight serial recognizer and in the meanwhile keeps a competitive running speed (10.5FPS). 

Moreover, our ARTS-P also outperforms previous methods by a large margin, achieving an E2E F-measure of 75.8\% at 13.0 FPS. For a faster ARTS-P R18 version, we adopt ResNet18 as backbone but can still achieve much better E2E performance compared with ABCNet (73.5\% \vs\ 64.2\%) while keeping a comparable running speed (17.0 FPS \vs\ 17.9 FPS). For our real-time version, we achieve the fastest running speed of 28FPS with a competitive E2E F-measure of 65.9\%.

\begin{table*}[ht!]
% \vspace{-1mm}
\centering
\small
\begin{tabular}{|c|c|c|c|c|c|c|c|c|c|}\hline
\multirow{2}{*}{Method} & \multirow{2}{*}{Venue} & \multirow{2}{*}{Backbone} & \multicolumn{3}{c|}{Detection} & \multicolumn{3}{c|}{End-to-End} & \multirow{2}{*}{FPS} \\
\cline{4-9} & & & Precision & Recall & F-measure & Strong & Weak & Generic & \\ \hline
MaskTextspotter v1~\cite{2018eccv_lyu_masktextspotterv1} & ECCV'18 & ResNet50 & 91.6 & 81.0 & 86.0 & 79.3 & 73.0 & 62.4 & 4.8\\
FOTS~\cite{2018cvpr_liu_fots} & CVPR'18 & ResNet50 & 91.0 & 85.2 & 88.0 & 81.1 & 75.9 & 60.8 & 7.8\\
He \etal~\cite{2018cvpr_he_e2etextspotter} & CVPR'18 & PVA & 87.0 & 86.0 & 87.0 & 82.0 & 77.0 & 63.0 & -\\
CharNet R-50~\cite{2019iccv_xing_charnet} & ICCV'19 & ResNet50 & 91.2 & \textbf{88.3} & \textbf{89.7} & 80.1 & 74.5 & 62.2 & -\\
TextDragon~\cite{2019iccv_feng_textdragonv1} & ICCV'19 & VGG16 & \textbf{92.5} & 83.8 & 87.9 & \textbf{82.5} & \textbf{78.3} & 65.2 & -\\
% Unconstrained~\cite{2019iccv_qin_unconstrained} & ICCV'19 & 89.4 & 85.8 & 87.5 & 83.4 & 79.9 & 67.9 & 4.8\\ 
TextPerceptron~\cite{2020aaai_qiao_textperceptron} & AAAI'20 & ResNet50 & 92.3 & 82.5 & 87.1 & 80.5 & 76.6 & 65.1 & -\\
Boundary~\cite{2020aaai_wang_boundary} & AAAI'20 & ResNet50 & 89.8 & 87.5 & 88.6 & 79.7 & 75.2 & 64.1 & -\\
\hline
ARTS-P (ours) & - & ResNet50 & 88.9 & 87.3 & 88.2 & 80.6 & 76.8 & 66.6 & \textbf{12.0}\\
ARTS-S (ours) & - & ResNet50 & 90.7 & 86.1 & 88.3 & 81.5 & 77.3 & \textbf{68.7} & 10.0\\
\hline
\end{tabular}
% \vspace{-3mm}
\caption{Quantitative results on ICDAR2015~\cite{2015icdar_ic15}. "Strong", "Weak" and "Generic" indicate results with strong, weak and generic lexicon.}\label{tab:results_ic15}
\end{table*}

\subsubsection{Long Arbitrarily-Shaped Text Spotting.}
To verify the robustness of our method on long curved text, we also conduct experiments on a representative benchmark dataset called CTW1500. As can be seen in Table~\ref{tab:results_ctw}, our method can achieve highly competitive results both in end-to-end text spotting metric and detection metric. Specifically, our proposed network, as a regression-based method, can achieve better results in 
E2E metric even compared with those state-of-the-art segmentation-based methods~\cite{2020aaai_qiao_textperceptron} (60.6\% \vs\ 57.0\%). When compared with previous regression-based methods (\eg, ABCNet~\cite{2020cvpr_liu_abcnet}), our method achieves a even larger advantage (60.6\% \vs\ 45.2\%). The results demonstrate that our network, even as a regression-based method, is still robust to those extremely long curved text instances which could be very difficult for previous regression-based methods due to the extreme aspect ratios.

\begin{table}[ht!]
% \vspace{-1mm}
\centering
\small
\begin{tabular}{|c|c|c|c|}\hline
\multirow{2}{*}{Method} & Detection & \multicolumn{2}{c|}{End-to-End}\\
\cline{2-4} & F-measure & None & Full\\
\hline
TextSnake*~\cite{2018eccv_long_textsnake} & 75.6 & - & -\\
PSENet~\cite{2019cvpr_wang_psenet} & 82.2 & - & -\\
FOTS*~\cite{2018cvpr_liu_fots} & 62.8 & 21.1 & 39.7 \\
TextDragon*~\cite{2019iccv_feng_textdragonv1} & 83.6 & 39.7 & 72.4 \\
ABCNet~\cite{2020cvpr_liu_abcnet} & - & 45.2 & 74.1 \\
TextPerceptron~\cite{2020aaai_qiao_textperceptron} & 84.6 & 57.0 & - \\
\hline
ARTS (ours) & \textbf{84.9} & \textbf{60.6} & \textbf{80.4}\\
\hline
\end{tabular}
% \vspace{-0.3cm}
\caption{Quantitative results on CTW1500. * indicates results are from ~\cite{2019iccv_feng_textdragonv1}. None and Full indicate using no lexicon and full lexicon, respectively.}\label{tab:results_ctw}
\end{table}

\subsubsection{Multi-Oriented Text Spotting.}
Though our method mainly focuses on arbitrarily-shaped text spotting, we can still achieve state-of-the-art performance on multi-oriented dataset ICDAR2015. As can be seen in Table~\ref{tab:results_ic15}, our method can surpass most of the previous state-of-the-art methods while keeping the fastest running speed. Specifically, our ARTS-S achieves the highest E2E F-measure of 68.7\% with generic lexicon and runs at a fast running speed (10.0FPS). And our ARTS-P can still achieves a competitive E2E F-measure of 66.6 with generic lexicon at 12.0 FPS, which is 50\% faster than previous methods.

\subsection{Ablation Study}

\subsubsection{Comparisons with BezierAlign.}
We have theoretically emphasized the advantages of our ARM in the above section. Here we conduct experiments on Total-Text to show the performance differences between our ARM and a representative SOTA extracting method BezierAlign~\cite{2020cvpr_liu_abcnet}. In our experiments, we directly use the BezierAlign operator to replace our ARM as feature extraction module in our pipeline and follow the training strategy provided by the official code repository (\#1). We fix all the other settings including decoder architectures and data augmentations.

As shown in Table~\ref{tab:ablation_comparison}, using BezierAlign (\#1) suffers a big performance drop (74.9\% \vs\ 77.1\%) compared with using our differentiable ARM (\#2). Though model with BezierAlign can produce smoother text features, its detection branch loses the ability to learn from recognition information and eventually leads to performance drop, indicating that using our proposed ARM to deal with the ``inconsistency problem'' 
% do have great importance.
% do have a point.
are quite essential for robust text spotting.
% this change will lead to a big performance drop.

Note that BezierAlign with our architecture achieves better results than the original ABCNet due to the using of attention-based recognizer and different data augmentations.

% The results are shown in Table.\ref{}

\begin{table}[ht!]
\centering
\small
\begin{tabular}{|c|c|c|c|c|}\hline
\multirow{2}{*}{\#} & \multirow{2}{*}{FEM} & \multirow{2}{*}{Rec-BP} & \multicolumn{2}{c|}{Total-Text}\\
\cline{4-5} & & & Det-F & E2E-F\\
\hline
% Det-only & $\times$ & $\checkmark$ & $\times$ \\ 
1 & BezierAlign~\cite{2020cvpr_liu_abcnet} & $\times$ & 85.4 & 74.9\\
% \textbf{Ours-1} & $\checkmark$ & 0.2 & $\times$ & 86.3 & 76.7\\
2 & ARM (ours) & $\checkmark$ & \textbf{86.5} & \textbf{77.1} \\
% \#3 & $\checkmark$ & $\checkmark$ & \textbf{86.5} & \textbf{77.1}\\

% \multirow{4}{*}{ARTS-S} & w/o. rec loss bp & 85.03 & 71.76\\
% \cline{2-4} & w. rec loss bp & - & -\\
% \cline{2-4} & w. rec loss bp & - & -\\
% \cline{2-4} & det only & - & -\\
\hline
\end{tabular}
\caption{Performance comparison between BezierAlign and our ARM on Total-Text. FEM means the feature extraction module, and Rec-BP shows whether it can back-propagate recognition losses into detection branch.}\label{tab:ablation_comparison}
\end{table}

\subsubsection{Effectiveness of Back-Propagating Recognition Loss to Detection Branch.}
To validate the effectiveness of back-propagating recognition loss, we design three groups of ablation experiments on Total-Text. 
For the first group (\#1), we train recognition branch with features extracted by the ground-truth polygons and thus cut off the loss back-propagation.
%For the first group (\#1), we simulate previous methods just like what ~\cite{2018cvpr_liu_fots}\cite{2020cvpr_liu_abcnet} did by simply using the ground-truth polygon to extract features for training recognition branch.
For the second group (\#2), we use the predicted control points to rectify text features and enable recognition loss back-propagation, but in the meanwhile we set $\lambda_{rcp}$ to 0 so that RCPD head will only be optimized by recognition targets.
% For the third group(Ours-2), we use predicted control points, enable recognition loss back-propagation but set $\lambda_{rcp}$ to 0 so that our detection head will only be optimized by recognition targets.
As for the last group (\#3), we use predicted control points, enable recognition loss back-propagation and set $\lambda_{rcp}$ to 0.2 so that our RCPD head will be jointly optimized by detection and recognition targets.
Results can be seen in Table~\ref{tab:ablation_fully_e2e}.
With loss back-propagation, our method (\#3) outperforms the method without the ability (\#1) by +2.0\% in E2E F-measure and +1.4\% in detection F-measure, demonstrating the superiority of back-propagating recognition loss to detection branch. We also find a surprising result (\#2) that even without the supervision of control points targets, our network can still reach convergence and achieve good performance under the only supervision, \ie, recognition targets.

\begin{table}[ht!]
\centering
\small
\begin{tabular}{|c|cc|c|c|}\hline
\multirow{2}{*}{\#}  & \multirow{2}{*}{w/ ${\mathcal{L}}_{rcp}$} & \multirow{2}{*}{Rec-BP} & \multicolumn{2}{c|}{Total-Text}\\
\cline{4-5} & & & Det-F & E2E-F\\
\hline
% Det-only & $\times$ & $\checkmark$ & $\times$ \\ 
1 & $\checkmark$ & $\times$ & 85.1 & 75.1\\
% \textbf{Ours-1} & $\checkmark$ & 0.2 & $\times$ & 86.3 & 76.7\\
2 & $\times$ & $\checkmark$ & 86.1 & 77.0 \\
3 & $\checkmark$ & $\checkmark$ & \textbf{86.5} & \textbf{77.1}\\

% \multirow{4}{*}{ARTS-S} & w/o. rec loss bp & 85.03 & 71.76\\
% \cline{2-4} & w. rec loss bp & - & -\\
% \cline{2-4} & w. rec loss bp & - & -\\
% \cline{2-4} & det only & - & -\\
\hline
\end{tabular}
\caption{The Effectiveness of back-propagating recognition loss to detection branch. "w/ ${\mathcal{L}}_{rcp}$" means whether to use control points target as supervision and "Rec-BP" means whether to conduct recognition loss back-propagation.}\label{tab:ablation_fully_e2e}
\end{table}

\subsection{Visualization and Time Analysis}
\subsubsection{Result Visualization.} 
\begin{figure}[ht]
\begin{center}
% \fbox{\rule{0pt}{2in} \rule{.9\linewidth}{0pt}}
    \includegraphics[width=0.9\linewidth]{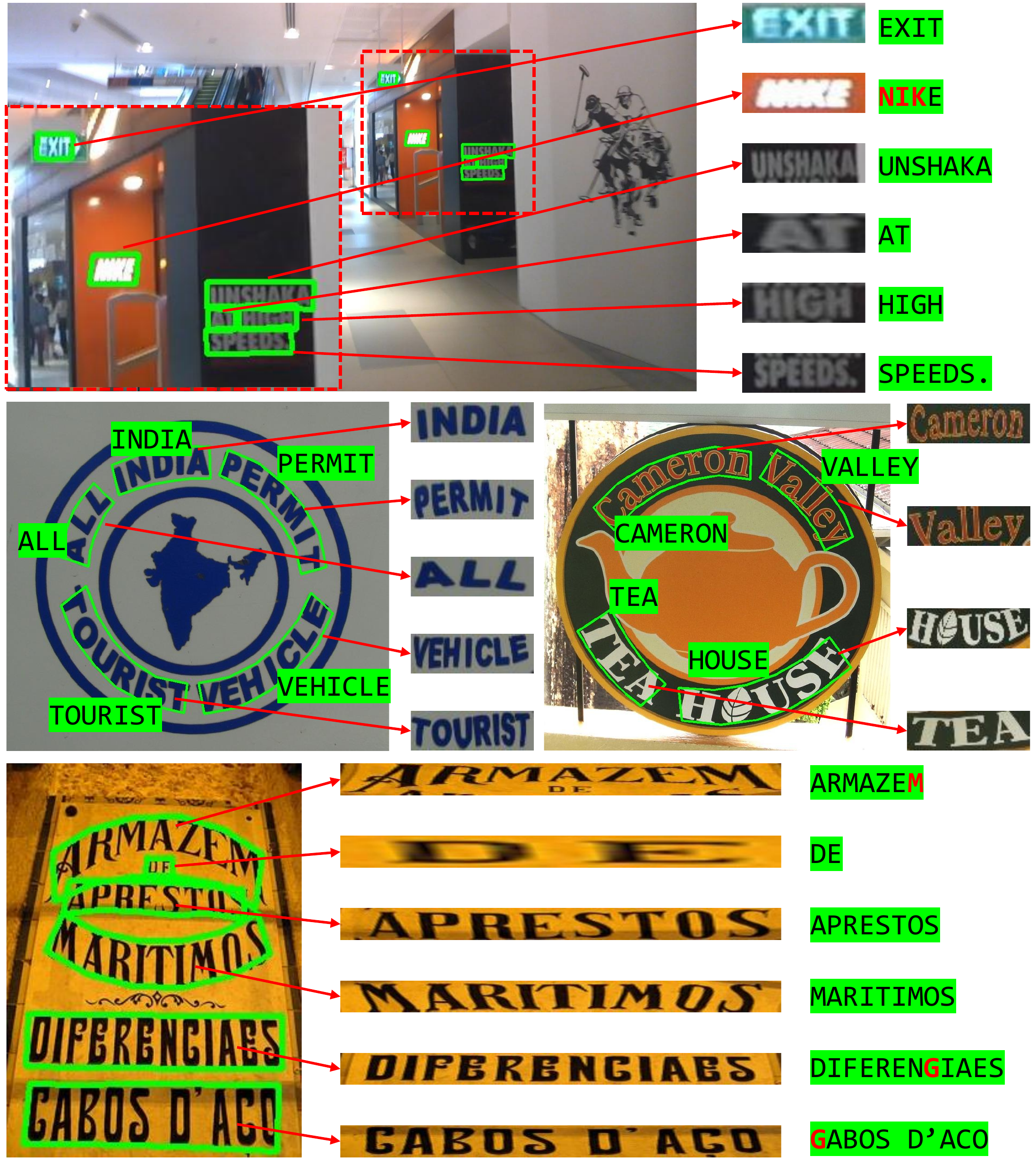}
\end{center}
% \vspace{-0.3cm}
\caption{Qualitative results of our method on ICDAR2015 (top), Total-Text (mid) and CTW1500 (bottom).}
\label{fig:vis}
% \vspace{-0.3cm}
\end{figure}

Qualitative results are illustrated in Figure~\ref{fig:vis}. Our proposed network can handle arbitrarily-shaped texts and seamlessly rectify them into straight texts for better recognition.

\subsubsection{Time Cost Analysis.} We analyze the time consumption of different components on Total-Text. All the experiments follow the same training protocol. As can be seen in Table~\ref{tab:ablation_recognizer}, using parallel instead of serial recognizer can reduce time cost for recognition to $30\%$ (6.9ms \vs\ 24.0ms) with only limited performance drop, making recognition branch a time-saving component and removing the barrier towards real-time scene text spotting.

\begin{table}[ht!]
% \vspace{-1mm}
\centering
\small
\begin{tabular}{|c|c|c|c|c|c|}\hline
\multirow{2}{*}{Method} & \multicolumn{4}{c|}{Time Cost (ms)} & \multirow{2}{*}{FPS}\\
\cline{2-5} & Backbone & Det & ARM & Rec & \\
\hline
ARTS-S (R50) & 8.1 & 54.0 & 4.6 & 24.0 & 10.5\\
ARTS-P (R50) & 8.1 & 54.0 & 4.6 & 6.9 & 13.0\\
ARTS-P (R18) & 4.5 & 39.0 & 4.7 & 6.9 & 17.0\\
% ARTS-RT & \textbf{4.4} & \textbf{24.4} & \textbf{4.2} & \textbf{6.6} & \textbf{23.8}\\
ARTS-RT (R18) & \textbf{4.0} & \textbf{19.2} & \textbf{4.3} & \textbf{6.8} & \textbf{28.0}\\
% \multirow{2}{*}{ARTS} & Serial & 73.5 & 10.5\\
% \cline{2-4} & Parallel & - & -\\
% \cline{2-4} & Parallel & - & -\\
% \cline{2-4} & (8, 128) & - & -\\
% \cline{2-4} & (16, 64) & - & -\\
\hline
\end{tabular}
\caption{Time analysis of different components on Total-Text. R* means ResNet with different layers. "Det" means detection and "Rec" means recognition.}\label{tab:ablation_recognizer}
% \vspace{-3mm}
\end{table}

\section{Conclusion}

In this paper, we systematically analyze the inconsistency between text detection and recognition. To tackle this problem, we design a differentiable auto-rectification module (ARM) together with a new training strategy to allow loss back-propagation from recognition branch to detection branch so that our detection branch can be jointly optimized by detection and recognition targets, thus largely alleviating the inconsistency problem.
Based on these, we propose a new arbitrarily-shaped text spotter, termed ARTS, to fast detect and recognize scene texts.
Extensive experiments on both arbitrarily-shaped (Total-Text and CTW1500) and multi-oriented (ICDAR2015) benchmark datasets demonstrate that our proposed ARTS can achieve state-of-the-art performance in terms of both accuracy and efficiency.

\bibliography{aaai22.bib}

\begin{thebibliography}{26}
\providecommand{\natexlab}[1]{#1}

\bibitem[{Baek et~al.(2019)Baek, Kim, Lee, Park, Han, Yun, Oh, and
  Lee}]{2019iccv_baek_whatiswrong}
Baek, J.; Kim, G.; Lee, J.; Park, S.; Han, D.; Yun, S.; Oh, S.~J.; and Lee, H.
  2019.
\newblock What Is Wrong With Scene Text Recognition Model Comparisons? Dataset
  and Model Analysis.
\newblock In \emph{Proceedings of the IEEE/CVF International Conference on
  Computer Vision (ICCV)}.

\bibitem[{Busta, Neumann, and Matas(2017)}]{2017iccv_busta_deeptextspotter}
Busta, M.; Neumann, L.; and Matas, J. 2017.
\newblock Deep TextSpotter: An End-To-End Trainable Scene Text Localization and
  Recognition Framework.
\newblock In \emph{Proceedings of the IEEE International Conference on Computer
  Vision (ICCV)}.

\bibitem[{{Ch'ng} and {Chan}(2017)}]{2017icdar_totaltext}
{Ch'ng}, C.~K.; and {Chan}, C.~S. 2017.
\newblock Total-Text: A Comprehensive Dataset for Scene Text Detection and
  Recognition.
\newblock In \emph{2017 14th IAPR International Conference on Document Analysis
  and Recognition (ICDAR)}, volume~01, 935--942.

\bibitem[{Feng et~al.(2019)Feng, He, Yin, Zhang, and
  Liu}]{2019iccv_feng_textdragonv1}
Feng, W.; He, W.; Yin, F.; Zhang, X.-Y.; and Liu, C.-L. 2019.
\newblock TextDragon: An End-to-End Framework for Arbitrary Shaped Text
  Spotting.
\newblock In \emph{Proceedings of the IEEE/CVF International Conference on
  Computer Vision (ICCV)}.

\bibitem[{Girshick(2015)}]{2015iccv_girshick_fastrcnn}
Girshick, R. 2015.
\newblock Fast R-CNN.
\newblock In \emph{Proceedings of the IEEE International Conference on Computer
  Vision (ICCV)}.

\bibitem[{He et~al.(2017)He, Gkioxari, Dollar, and
  Girshick}]{2017iccv_he_maskrcnn}
He, K.; Gkioxari, G.; Dollar, P.; and Girshick, R. 2017.
\newblock Mask R-CNN.
\newblock In \emph{Proceedings of the IEEE International Conference on Computer
  Vision (ICCV)}.

\bibitem[{He et~al.(2016)He, Zhang, Ren, and Sun}]{2016cvpr_he_resnet}
He, K.; Zhang, X.; Ren, S.; and Sun, J. 2016.
\newblock Deep Residual Learning for Image Recognition.
\newblock In \emph{Proceedings of the IEEE Conference on Computer Vision and
  Pattern Recognition (CVPR)}.

\bibitem[{He et~al.(2018)He, Tian, Huang, Shen, Qiao, and
  Sun}]{2018cvpr_he_e2etextspotter}
He, T.; Tian, Z.; Huang, W.; Shen, C.; Qiao, Y.; and Sun, C. 2018.
\newblock An End-to-End TextSpotter With Explicit Alignment and Attention.
\newblock In \emph{Proceedings of the IEEE Conference on Computer Vision and
  Pattern Recognition (CVPR)}.

\bibitem[{Jaderberg et~al.(2015)Jaderberg, Simonyan, Zisserman, and
  Kavukcuoglu}]{2015nips_stn}
Jaderberg, M.; Simonyan, K.; Zisserman, A.; and Kavukcuoglu, K. 2015.
\newblock Spatial Transformer Networks.
\newblock In \emph{Proceedings of the 28th International Conference on Neural
  Information Processing Systems - Volume 2}, NIPS'15, 2017–2025. Cambridge,
  MA, USA: MIT Press.

\bibitem[{{Karatzas} et~al.(2015){Karatzas}, {Gomez-Bigorda}, {Nicolaou},
  {Ghosh}, {Bagdanov}, {Iwamura}, {Matas}, {Neumann}, {Chandrasekhar}, {Lu},
  {Shafait}, {Uchida}, and {Valveny}}]{2015icdar_ic15}
{Karatzas}, D.; {Gomez-Bigorda}, L.; {Nicolaou}, A.; {Ghosh}, S.; {Bagdanov},
  A.; {Iwamura}, M.; {Matas}, J.; {Neumann}, L.; {Chandrasekhar}, V.~R.; {Lu},
  S.; {Shafait}, F.; {Uchida}, S.; and {Valveny}, E. 2015.
\newblock ICDAR 2015 competition on Robust Reading.
\newblock In \emph{2015 13th International Conference on Document Analysis and
  Recognition (ICDAR)}, 1156--1160.

\bibitem[{Liao et~al.(2021)Liao, Lyu, He, Yao, Wu, and
  Bai}]{2019pami_liao_masktextspotterv2}
Liao, M.; Lyu, P.; He, M.; Yao, C.; Wu, W.; and Bai, X. 2021.
\newblock Mask TextSpotter: An End-to-End Trainable Neural Network for Spotting
  Text with Arbitrary Shapes.
\newblock \emph{IEEE transactions on pattern analysis and machine
  intelligence}, 43(2): 532—548.

\bibitem[{Liao et~al.(2020)Liao, Pang, Huang, Hassner, and
  Bai}]{2020eccv_liao_masktextspotterv3}
Liao, M.; Pang, G.; Huang, J.; Hassner, T.; and Bai, X. 2020.
\newblock Mask TextSpotter v3: Segmentation Proposal Network for Robust Scene
  Text Spotting.
\newblock In Vedaldi, A.; Bischof, H.; Brox, T.; and Frahm, J.-M., eds.,
  \emph{Computer Vision -- ECCV 2020}, 706--722. Cham: Springer International
  Publishing.

\bibitem[{Lin et~al.(2017)Lin, Dollar, Girshick, He, Hariharan, and
  Belongie}]{2017cvpr_lin_fpn}
Lin, T.-Y.; Dollar, P.; Girshick, R.; He, K.; Hariharan, B.; and Belongie, S.
  2017.
\newblock Feature Pyramid Networks for Object Detection.
\newblock In \emph{Proceedings of the IEEE Conference on Computer Vision and
  Pattern Recognition (CVPR)}.

\bibitem[{Liu et~al.(2018)Liu, Liang, Yan, Chen, Qiao, and
  Yan}]{2018cvpr_liu_fots}
Liu, X.; Liang, D.; Yan, S.; Chen, D.; Qiao, Y.; and Yan, J. 2018.
\newblock FOTS: Fast Oriented Text Spotting With a Unified Network.
\newblock In \emph{Proceedings of the IEEE Conference on Computer Vision and
  Pattern Recognition (CVPR)}.

\bibitem[{Liu et~al.(2020)Liu, Chen, Shen, He, Jin, and
  Wang}]{2020cvpr_liu_abcnet}
Liu, Y.; Chen, H.; Shen, C.; He, T.; Jin, L.; and Wang, L. 2020.
\newblock ABCNet: Real-Time Scene Text Spotting With Adaptive Bezier-Curve
  Network.
\newblock In \emph{IEEE/CVF Conference on Computer Vision and Pattern
  Recognition (CVPR)}.

\bibitem[{Long et~al.(2018)Long, Ruan, Zhang, He, Wu, and
  Yao}]{2018eccv_long_textsnake}
Long, S.; Ruan, J.; Zhang, W.; He, X.; Wu, W.; and Yao, C. 2018.
\newblock TextSnake: A Flexible Representation for Detecting Text of Arbitrary
  Shapes.
\newblock In \emph{Proceedings of the European Conference on Computer Vision
  (ECCV)}.

\bibitem[{Lyu et~al.(2018)Lyu, Liao, Yao, Wu, and
  Bai}]{2018eccv_lyu_masktextspotterv1}
Lyu, P.; Liao, M.; Yao, C.; Wu, W.; and Bai, X. 2018.
\newblock Mask TextSpotter: An End-to-End Trainable Neural Network for Spotting
  Text with Arbitrary Shapes.
\newblock In \emph{Proceedings of the European Conference on Computer Vision
  (ECCV)}.

\bibitem[{{Nayef} et~al.(2017){Nayef}, {Yin}, {Bizid}, {Choi}, {Feng},
  {Karatzas}, {Luo}, {Pal}, {Rigaud}, {Chazalon}, {Khlif}, {Luqman}, {Burie},
  {Liu}, and {Ogier}}]{2017icdar_mlt17}
{Nayef}, N.; {Yin}, F.; {Bizid}, I.; {Choi}, H.; {Feng}, Y.; {Karatzas}, D.;
  {Luo}, Z.; {Pal}, U.; {Rigaud}, C.; {Chazalon}, J.; {Khlif}, W.; {Luqman},
  M.~M.; {Burie}, J.; {Liu}, C.; and {Ogier}, J. 2017.
\newblock ICDAR2017 Robust Reading Challenge on Multi-Lingual Scene Text
  Detection and Script Identification - RRC-MLT.
\newblock In \emph{2017 14th IAPR International Conference on Document Analysis
  and Recognition (ICDAR)}, volume~01, 1454--1459.

\bibitem[{Qiao et~al.(2020)Qiao, Tang, Cheng, Xu, Niu, Pu, and
  Wu}]{2020aaai_qiao_textperceptron}
Qiao, L.; Tang, S.; Cheng, Z.; Xu, Y.; Niu, Y.; Pu, S.; and Wu, F. 2020.
\newblock Text Perceptron: Towards End-to-End Arbitrary-Shaped Text Spotting.
\newblock \emph{Proceedings of the AAAI Conference on Artificial Intelligence},
  34(07): 11899--11907.

\bibitem[{Qin et~al.(2019)Qin, Bissacco, Raptis, Fujii, and
  Xiao}]{2019iccv_qin_unconstrained}
Qin, S.; Bissacco, A.; Raptis, M.; Fujii, Y.; and Xiao, Y. 2019.
\newblock Towards Unconstrained End-to-End Text Spotting.
\newblock In \emph{Proceedings of the IEEE/CVF International Conference on
  Computer Vision (ICCV)}.

\bibitem[{{Shi} et~al.(2019){Shi}, {Yang}, {Wang}, {Lyu}, {Yao}, and
  {Bai}}]{2018pami_shi_aster}
{Shi}, B.; {Yang}, M.; {Wang}, X.; {Lyu}, P.; {Yao}, C.; and {Bai}, X. 2019.
\newblock ASTER: An Attentional Scene Text Recognizer with Flexible
  Rectification.
\newblock \emph{IEEE Transactions on Pattern Analysis and Machine
  Intelligence}, 41(9): 2035--2048.

\bibitem[{Tian et~al.(2019)Tian, Shen, Chen, and He}]{2019iccv_tian_fcos}
Tian, Z.; Shen, C.; Chen, H.; and He, T. 2019.
\newblock FCOS: Fully Convolutional One-Stage Object Detection.
\newblock In \emph{Proceedings of the IEEE/CVF International Conference on
  Computer Vision (ICCV)}.

\bibitem[{Wang et~al.(2020)Wang, Lu, Zhang, Yang, Bai, Xu, He, Wang, and
  Liu}]{2020aaai_wang_boundary}
Wang, H.; Lu, P.; Zhang, H.; Yang, M.; Bai, X.; Xu, Y.; He, M.; Wang, Y.; and
  Liu, W. 2020.
\newblock All You Need Is Boundary: Toward Arbitrary-Shaped Text Spotting.
\newblock \emph{Proceedings of the AAAI Conference on Artificial Intelligence},
  34(07): 12160--12167.

\bibitem[{Wang et~al.(2019)Wang, Xie, Li, Hou, Lu, Yu, and
  Shao}]{2019cvpr_wang_psenet}
Wang, W.; Xie, E.; Li, X.; Hou, W.; Lu, T.; Yu, G.; and Shao, S. 2019.
\newblock Shape Robust Text Detection With Progressive Scale Expansion Network.
\newblock In \emph{Proceedings of the IEEE/CVF Conference on Computer Vision
  and Pattern Recognition (CVPR)}.

\bibitem[{Xing et~al.(2019)Xing, Tian, Huang, and
  Scott}]{2019iccv_xing_charnet}
Xing, L.; Tian, Z.; Huang, W.; and Scott, M.~R. 2019.
\newblock Convolutional Character Networks.
\newblock In \emph{Proceedings of the IEEE/CVF International Conference on
  Computer Vision (ICCV)}.

\bibitem[{Yuliang et~al.(2017)Yuliang, Lianwen, Shuaitao, and
  Sheng}]{2017arxiv_liu_ctw1500}
Yuliang, L.; Lianwen, J.; Shuaitao, Z.; and Sheng, Z. 2017.
\newblock Detecting Curve Text in the Wild: New Dataset and New Solution.
\newblock arXiv:1712.02170.

\end{thebibliography}

\end{document}